\documentclass{vgtc}                          

\ifpdf
  \pdfoutput=1\relax                   
  \usepackage{graphicx}                
  \DeclareGraphicsExtensions{.pdf,.png,.jpg,.jpeg} 
\else
  \usepackage{graphicx}                
  \DeclareGraphicsExtensions{.eps}     
\fi%

\graphicspath{{figures/}{pictures/}{images/}{./}} 
\usepackage{tcolorbox}
\usepackage{pdfpages}

\usepackage[font=scriptsize]{subcaption}

\usepackage{microtype}                 
\PassOptionsToPackage{warn}{textcomp}  
\usepackage{textcomp}                  
\usepackage{mathptmx}                  
\usepackage{times}                     
\usepackage{cite}                      
\usepackage{tabu}                      
\usepackage{booktabs}                  
\usepackage{color}
\definecolor{mygray}{rgb}{0.466327, 0.466327, 0.466327}

\usepackage{multicol}
\onlineid{0}

\vgtccategory{Research}

\title{Perceptually Optimized Color Selection for Visualization}

\author{Subhrajyoti Maji\thanks{e-mail: majis@tcd.ie} %
\and John Dingliana\thanks{e-mail: dinglijl@tcd.ie} %
}
\affiliation{\scriptsize Graphics Vision and Visualization Group \\Trinity College Dublin}

\abstract{
We propose an approach, called the \emph{Equilibrium Distribution Model} (\emph{EDM}), for automatically selecting colors with optimum perceptual contrast for scientific visualization. Given any number of features that need to be emphasized in a visualization task, our approach derives evenly distributed points in the \emph{CIELAB} color space to assign colors to the features so that the minimum Euclidean Distance among the colors are optimized. Our approach can assign colors with high perceptual contrast even for very high numbers of features, where other color selection methods typically fail. 
We compare our approach with the widely used \emph{Harmonic} color selection scheme and demonstrate that while the harmonic scheme can achieve reasonable color contrast for visualizing up to $20$ different features, our \emph{Equilibrium} scheme provides significantly better contrast and achieves perceptible contrast for visualizing even up to $100$ unique features.

} 

\CCScatlist{
  \CCScatTwelve{Color}{Visu\-al\-iza\-tion}{Visu\-al\-iza\-tion techniques};
  \CCScatTwelve{Human-centered computing}{Visu\-al\-iza\-tion}{Visualization design and evaluation methods}{}
}

\begin{document}


\firstsection{Introduction}
\maketitle

Mapping perceptually distinct colors to different aspects of a data is an integral part of scientific visualization. Therefore, the choice of colors can make a significant difference in gaining insight from of a given visualization. Common practice ranges from a manual selection of colors, use of some standard recurring palettes or use of palette design tool such as those in \emph{ColorBrewer} \cite{harrower2003colorbrewer} or \emph{PRAVDAColor} \cite{bergman1995rule}. Such practices work reasonably well if the number of features to be visualized is relatively small. However, they become cumbersome and ineffective if we wish to visualize data comprising an unusually high number of features of the order of $50-100$. In such cases, there is a need to select the set of colors based on some criteria automatically. \emph{Color harmonics} and \emph{color opponency} are some of the popular choices for the practical design of colormaps. However, these approaches do not fully exploit the whole range of a perceptual color space, and hence the perceptual contrast of the resulting colors decreases rapidly with increasing the number of colors. A distribution of $20$ different colors in the \emph{CIELAB} color space using \emph{harmonic} color selection scheme is shown in Figure \ref{fig:chromaticityharmonic}.

Our solution aims to utilize the maximum range in the \emph{CIELAB} color space by evenly distributing the color points within the space. We demonstrate the effectiveness of using this approach over the \emph{harmonic} color selection scheme in visualizing a 3D volume dataset and a 2D data visualization. A measure of perceptual contrast between colors indicates that our approach outperforms the harmonic color scheme and leads to distinct colors with contrast well above the \emph{Just Noticeable Difference (JND)} threshold even for high numbers of unique independent features.

\section{Methodology}

\begin{figure}[h]
\centering
 \begin{subfigure}[t]{0.45\linewidth}
 \tcbox[colback=white]{\includegraphics[width=0.7\linewidth]{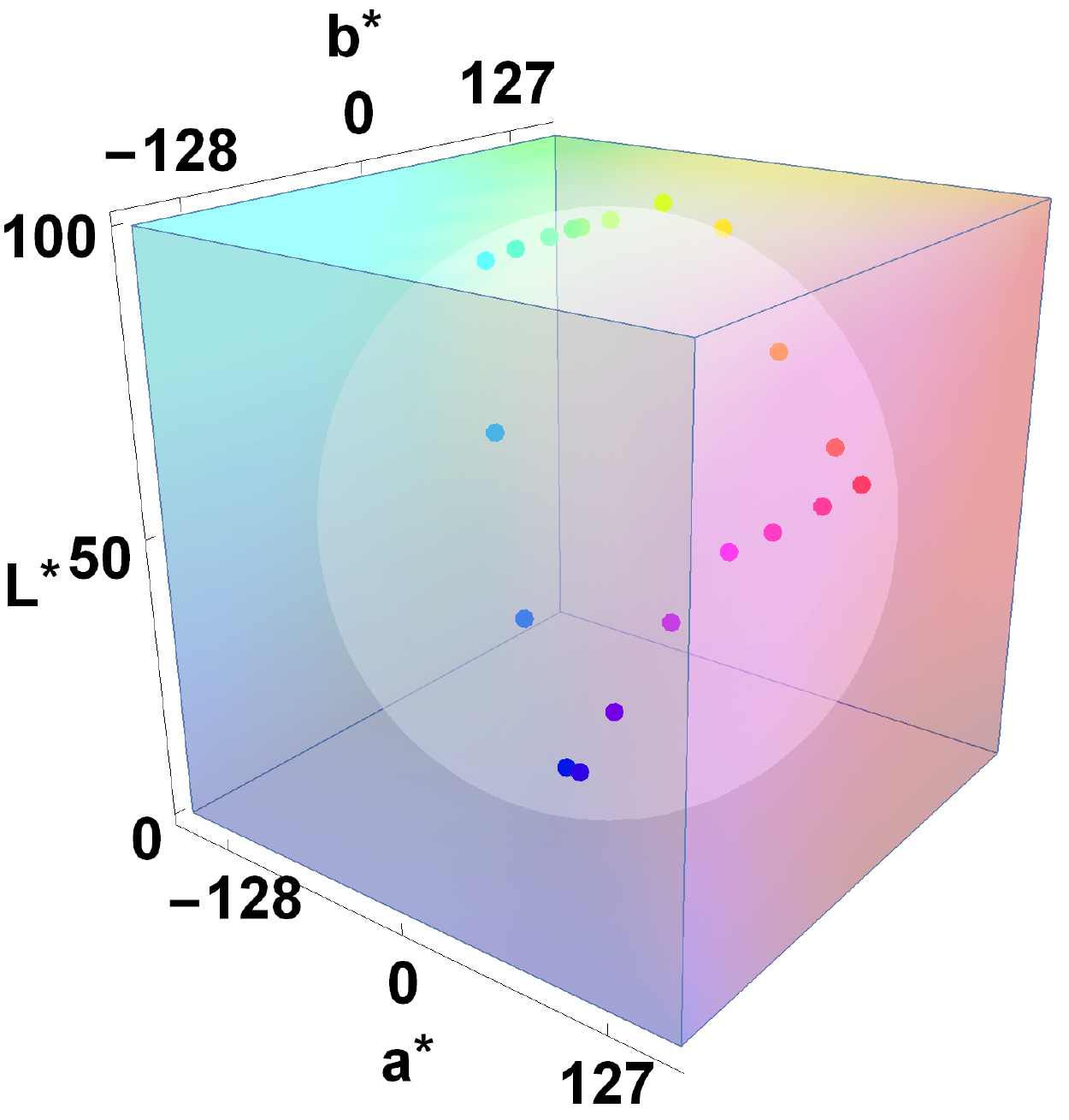}}
   \vspace{-.6\baselineskip}
   \caption{Harmonic color scheme}
 \label{fig:chromaticityharmonic} 
 \end{subfigure} 
\quad 
 \begin{subfigure}[t]{0.45\linewidth}
  \tcbox[colback=white]{\includegraphics[width=0.7\linewidth]{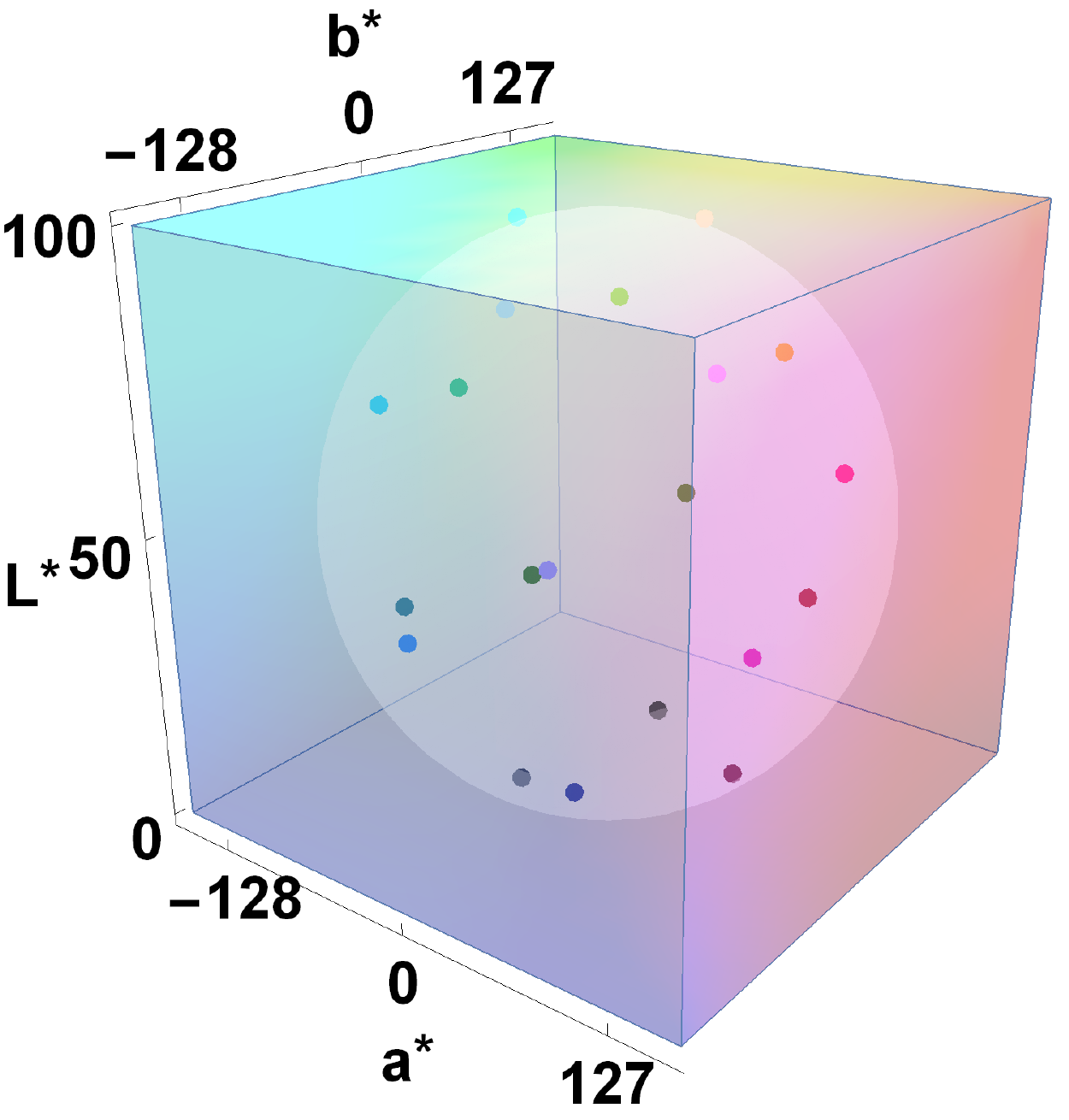}} 

   \vspace{-.6\baselineskip}
 \caption{Equilibrium color scheme}
  \label{fig:chromaticityEDM} 
 \end{subfigure}
 \vspace{-.4\baselineskip}
 \caption{Equilibrium distribution of $20$ colors in \emph{CIELAB} color space using (a) \emph{Harmonic} and (b) \emph{Equilibrium} color schemes.}
\label{fig:chromaticityplot}
\end{figure}

The human visual system tends to not perceive absolute colors of objects but rather the differences between them, i.e., the contrast. 
The most common type of color contrast is the contrast between the foreground color and the background color.  According to Itten \cite{itten1970elements}, the human response to color stimuli reaches an equilibrium state when it receives a signal from a mid-gray object. Based on this fact, the first objective of our algorithm is to select colors which are equidistant from the mid-gray value in a perceptual color space (\emph{CIELAB} in this case). This task can be accomplished by constructing a hypothetical sphere with center at the mid-gray value and spanning over the entire color space.  Now all the colors corresponding to the points lying on the surface of the sphere will create the same sensation when displayed as a foreground color against the gray background. The second task is to distribute these points on the spherical surface evenly. For this purpose, we consider the analogy of spherical distribution of charged particles in a static equilibrium state in which the net potential is to be minimized through moving individual charged particles. A demonstration of such an equilibrium distribution of $20$ different colors in the \emph{CIELAB} color space is shown in Figure \ref{fig:chromaticityEDM}. Moreover, to confirm that this technique distributes the points maximally distant apart, a comparison (see Table \ref{table:distancecomparison}) is done between the edge lengths of the Platonic solids with the minimum distances achieved for the corresponding number of vertices using this technique. It can be observed that \emph{EDM} matches perfectly with the edge lengths of the Platonic solids except for the Cube and Dodecahedron where the corresponding vertices in the equilibrium form different shapes other than these two. Although we specifically employ \emph{CIELAB} color space, the technique is flexible enough to select colors in any other perceptual color space.
    
%

\begin{table}[]
\caption{Comparison of minimum distances using \emph{EDM} with the edge lengths of Platonic solids. Asterisks (*) indicate cases where \emph{EDM} differs from the vertices of the corresponding Platonic solid.}
\label{table:distancecomparison}
\scriptsize
\centering
\begin{tabu}{l c c c}
\toprule
\multicolumn{1}{c}{\textbf{Platonic Solids}} & \multicolumn{1}{c}{\textbf{\begin{tabular}[c]{@{}c@{}}No. of\\  Vertices\end{tabular}}} & \multicolumn{1}{c}{\textbf{\begin{tabular}[c]{@{}c@{}}Edge\\  Length\end{tabular}}} & \multicolumn{1}{c}{\textbf{\begin{tabular}[c]{@{}c@{}}Min Distance\\  using \emph{EDM}\end{tabular}}} \\ \midrule

Tetrahedron                                    & 4                                                                                        & 1.63299                                                                              & 1.63299                                                                                                                \\ 
Octahedron                                     & 6                                                                                        & 1.41421                                                                              & 1.41421                                                                                                                \\ 
Cube                                           & 8                                                                                        & 1.1547                                                                               & 1.1712*                                                                                                                \\ 
Icosahedron                                    & 12                                                                                       & 1.05146                                                                              & 1.05146                                                                                                                \\ 
Dodecahedron                                   & 20                                                                                       & 0.713644                                                                             & 0.782961*                                                                                                              \\ \bottomrule
\end{tabu}

\end{table}

\section{Results}
\begin{figure}[h]
\centering
 \begin{subfigure}[t]{0.45\linewidth}
 \tcbox[colback=mygray]{\includegraphics[width=0.7\linewidth]{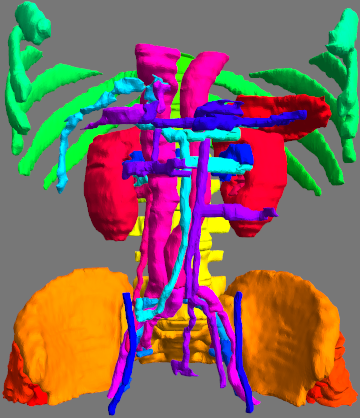}}
   \vspace{-.6\baselineskip}
   \caption{Harmonic color scheme}
 \label{fig:anatomyharmonic} 
 \end{subfigure} 
\quad 
 \begin{subfigure}[t]{0.45\linewidth}
  \tcbox[colback=mygray]{\includegraphics[width=0.7\linewidth]{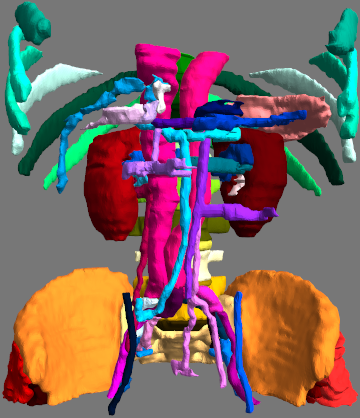}} 

   \vspace{-.6\baselineskip}
 \caption{Equilibrium color scheme}
  \label{fig:anatomyEDM} 
 \end{subfigure}
 \vspace{-.4\baselineskip}
 \caption{Visualization of a segmented anatomy data set with $75$ features using (a) \emph{Harmonic} and (b) \emph{Equilibrium} color schemes.}
\label{fig:anatomyvisualization}
\end{figure}

\begin{figure}[h]
\centering
 \begin{subfigure}[t]{0.45\linewidth}
 \centering
 \tcbox[colback=mygray]{\includegraphics[width=0.7\linewidth]{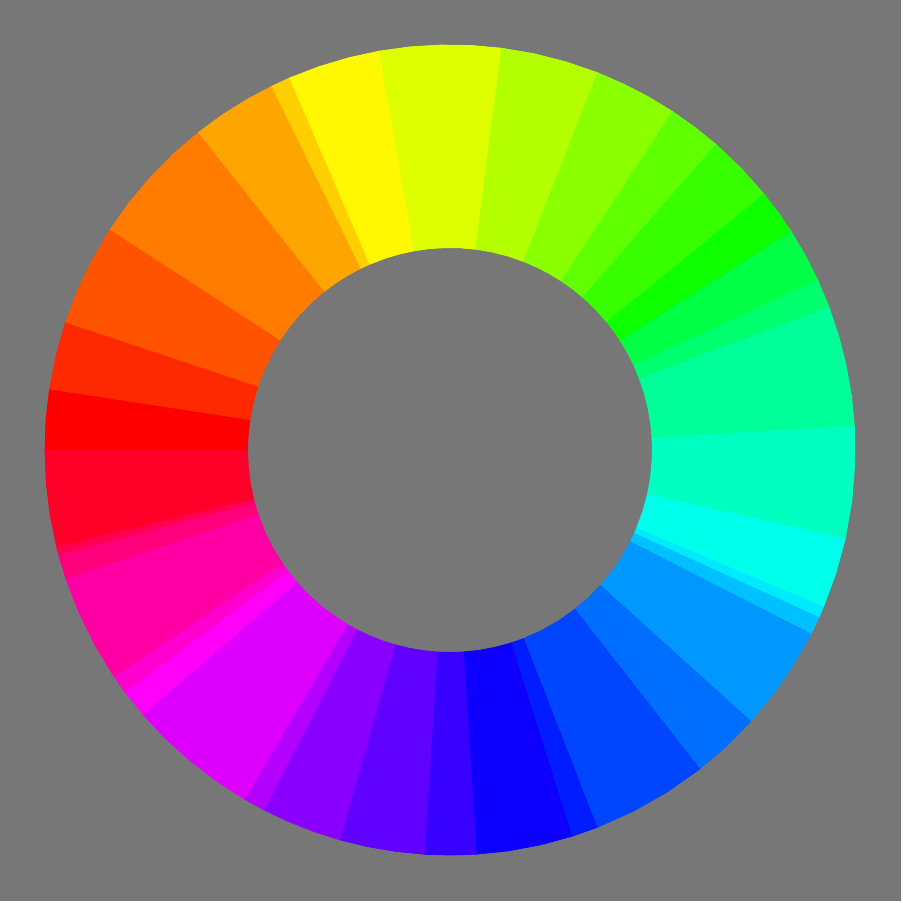}} 
\vspace{-.6\baselineskip}
\caption{Harmonic color scheme}
 \label{fig:pieharmonic} 
 \end{subfigure} 
\quad 
 \begin{subfigure}[t]{0.45\linewidth}
 \centering
  \tcbox[colback=mygray]{\includegraphics[width=0.7\linewidth]{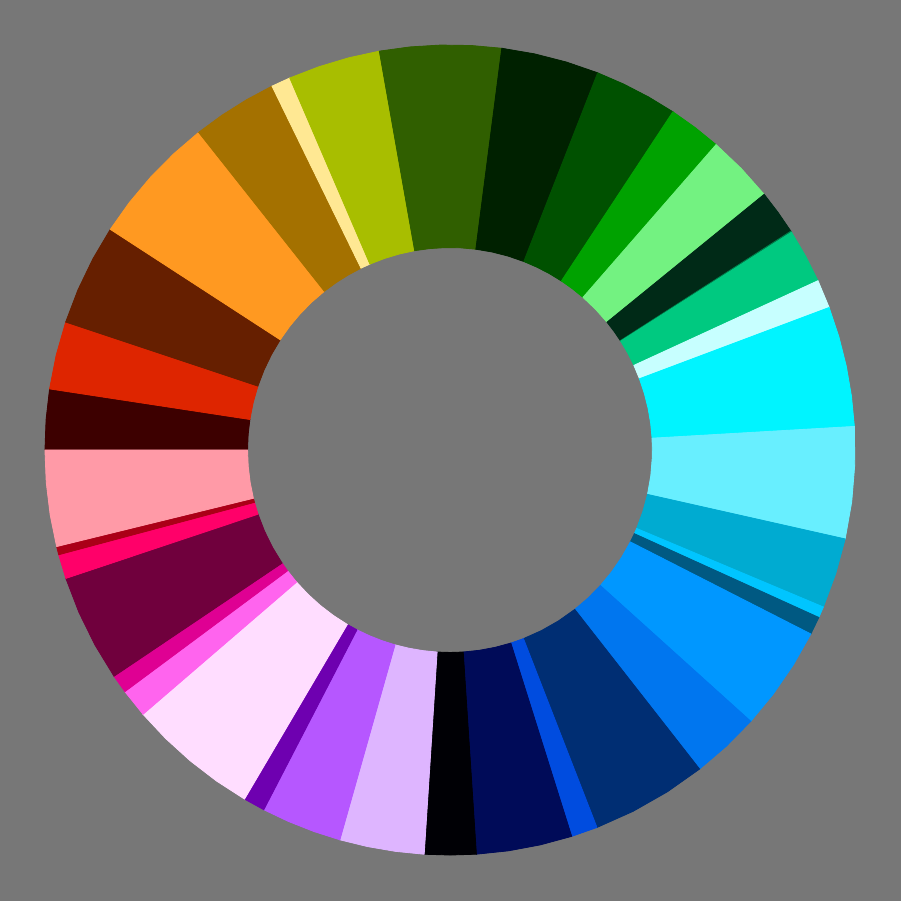}} 
  \vspace{-.6\baselineskip}
    \caption{Equilibrium color scheme}
  \label{fig:pieEDM} 
 \end{subfigure}
\caption{Visualization of a synthetic pie chart with $37$ segments using (a) \emph{Harmonic} and (b) \emph{Equilibrium} color schemes respectively.}
\label{fig:piechartvisualization}
\end{figure}

We demonstrate our approach by applying the colors to visualize a volume dataset (Figure \ref{fig:anatomyvisualization}) and a synthetic pie chart (Figure \ref{fig:piechartvisualization}) using \emph{equilibrium}  and \emph{harmonic} color selection schemes. The volume dataset~\cite{talos2015} comprises $94$ segments of a 3D CT scan and is rendered with a binary (fully visible or fully transparent) opacity value for each segment. Among these features, 19 features such as the muscles are made transparent to reveal the internal $75$ features in the visualization. By visual inspection, it is difficult to separate some of the green features in Figure \ref{fig:anatomyharmonic}, while the same features are more distinguishable in Figure \ref{fig:anatomyEDM}. Moreover, while all the colors in the \emph{harmonic} color scheme are highly saturated, a range of different saturations can be seen in the \emph{equilibrium} color scheme. Figure \ref{fig:piechartvisualization} shows the visualization of a pie chart, with $37$ features, using the same two color schemes. It can be readily observed that some segments are difficult to distinguish in the visualization generated using the \emph{harmonic} color scheme (Figure~\ref{fig:pieharmonic}). In comparison, the perceptual contrast is comparatively higher throughout when visualized using the \emph{equilibrium} color scheme (Figure \ref{fig:pieEDM}).


\section{Evaluation}

For objective evaluation, we measure the minimum contrast between colors selected by the algorithm. Here, we use two different measures of perceptual contrast. The first one, known as the \emph{CIE76 color distance}($\Delta E_{ab}^{*}$), considers perceptual contrast based on the Euclidean distance between two colors in the \emph{CIELAB} color space. To assess this contrast, we refer to the previous work by 
Mokrzycki and Tatol \cite{mokrzycki2011colour} who presented an experimentally verified statistical study, which determined that if the Euclidean Distance between two colors in the \emph{CIELAB} color space is greater than $5$ (where $L$ ranges from $0$ to $100$, $a$ ranges from $-128$ to $127$ and $b$ ranges from $-128$ to $127$) then the colors are easily distinguishable by a standard observer. Hence, the \emph{JND} corresponds to $\Delta E_{ab}^{*}\approx 5$. The second one is the most up-to-date measure of the perceptual contrast \cite{sharma2005ciede2000}, known as \emph{CIEDE2000 color distance}($\Delta E_{00}^{*}$) according to which the \emph{JND} corresponds to $\Delta E_{00}^{*}\approx 1$ \cite{yang2012color}.

\begin{figure}[h]
\centering
 \begin{subfigure}[t]{0.45\linewidth}
 \centering
 \tcbox[colback=white] {\includegraphics[width=0.7\linewidth]{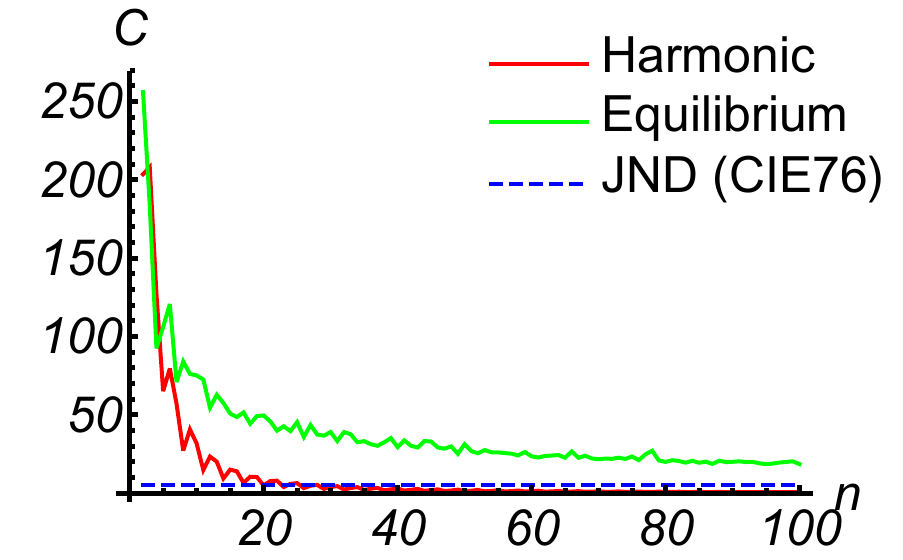}} 
   \vspace{-.6\baselineskip}
 \caption{CIE76 Color Distance}
 \label{fig:cie76} 
 \end{subfigure} 
\quad 
 \begin{subfigure}[t]{0.45\linewidth}
 \centering
  \tcbox[colback=white]{\includegraphics[width=0.7\linewidth]{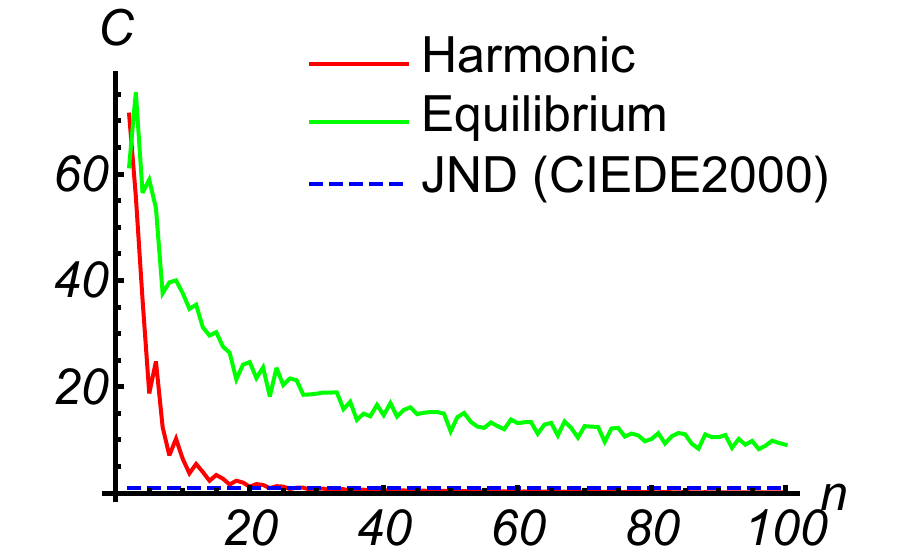}} 
    \vspace{-.6\baselineskip}
  \caption{CIEDE2000 Color Distance}
  \label{fig:cie2000} 
 \end{subfigure}
\caption{Comparison of perceptual contrasts achieved using \emph{Harmonic} and \emph{Equilibrium} color schemes with (a) \emph{CIE76} ($\emph{JND} =5$) and (b) \emph{CIEDE2000} ($\emph{JND} =1$) color distance formulae.}
\label{fig:perceptualcontrast}
\end{figure}

Figure \ref{fig:perceptualcontrast} shows a comparison of the contrast measures, $C$,  of the \emph{equilibrium} color scheme with the \emph{harmonic} color scheme for different numbers of unique features, $n$. 
We see from both the plots that the \emph{equilibrium} color scheme provides significantly better perceptual contrast. While the perceptual contrast achieved using the \emph{harmonic} color selection scheme reaches the \emph{JND} threshold at around $n=20$, the corresponding contrast using the \emph{equilibrium} color scheme remains significantly above the \emph{JND} even at $n=100$.


\section{Conclusion}

We have demonstrated the effectiveness of an automated selection of colors based on a novel \emph{equilibrium distribution model} and showed that the minimum perceptual contrast provided by this model is well above the threshold value for \emph{JND} even for a large number of unique features. We have further demonstrated its application in the volumetric visualization of a segmented CT scan and also in a simple data visualization. Although we have not considered the impacts of opacity and lighting in the final rendering, this is a problem that all other existing techniques would also face and we consider these as topics of investigation in the future work. We intend to carry out further experiments including other measures of contrast and also user-based studies.

\acknowledgments{
This research has been conducted with the financial support of Science
Foundation Ireland (SFI) under Grant Number 13/IA/1895.
}	

\bibliographystyle{abbrv-doi}

\bibliography{template}
\end{document}